\pdfoutput=1

\documentclass[11pt]{article}

\usepackage[]{acl}

\usepackage{times}
\usepackage{latexsym}
\usepackage{pgfplots}
\pgfplotsset{width=7.5cm,compat=1.9} 
\usetikzlibrary{patterns}
\usepackage{tikz}
\usepackage{enumitem}
\usepackage[T1]{fontenc}
\usepackage{xcolor,colortbl}
\usepackage{bm} 

\definecolor{RYB0}{RGB}{237,248,251}
\definecolor{RYB1}{RGB}{178,226,226}
\definecolor{RYB2}{RGB}{102,194,164}
\definecolor{RYB3}{RGB}{35,139,69}

\usepackage[utf8]{inputenc}

\usepackage{microtype}

\usepackage{inconsolata}

\usepackage{booktabs}
\usepackage{tabularx}
\usepackage{multirow}
\usepackage{siunitx}
\usepackage[normalem]{ulem}
\useunder{\uline}{\ul}{}
\usepackage{longtable}
\usetikzlibrary{arrows.meta}

\usepackage{graphicx}
\graphicspath{   {./images/} }

\usepgfplotslibrary{external}
\title{Using LLMs as Speech-to-Text Retrieval Systems}
\title{Transforming LLMs into Cross-modal and Cross-lingual Retrieval Systems}

\author{Frank Palma Gomez${}^{1}$ Ramon Sanabria${}^{2}$  Yun-hsuan Sung${}^{4}$\\ 
        {\bf Daniel Cer${}^{4}$} {\bf Siddharth Dalmia${}^{3{\ddagger}}$} {\bf Gustavo Hernandez Abrego${}^{4{\ddagger}}$}\\
      $^1$Boston University $^2$The University of Edinburgh $^3$Google DeepMind \\
      $^4$Google Research \\
      \texttt{fpg@bu.com}\thanks{${}$ Work done by Frank and Ramon during their internship in Google Research and Google DeepMind respectively.}\thanks{${}^{\ddagger}$ Equal Advising Contributions.}}
      
\begin{document}
\maketitle
\begin{abstract}

Large language models (LLMs) are trained on text-only data that go far beyond the languages with paired speech and text data. At the same time, Dual Encoder (DE) based retrieval systems project queries and documents into the same embedding space and have demonstrated their success in retrieval and bi-text mining.
To match speech and text in many languages, we propose using LLMs to initialize multi-modal DE retrieval systems. Unlike traditional methods, our system doesn't require speech data during LLM pre-training and can exploit LLM's multilingual text understanding capabilities to match speech and text in languages unseen during retrieval training.
Our multi-modal LLM-based retrieval system is capable of matching speech and text in 102 languages despite only training on 21 languages. 
Our system outperforms previous systems trained explicitly on all 102 languages.  We achieve a 10\% absolute improvement in Recall@1 averaged across these languages. 
Additionally, our model demonstrates cross-lingual speech and text matching, which is further enhanced by readily available machine translation data.

\end{abstract}

\section{Introduction}

LLMs have demonstrated their effectiveness in modelling textual sequences to tackle various downstream tasks \cite{brown2020language, hoffmann2022training, chowdhery2023palm}. 
This effectiveness has led to the development of powerful LLMs capable of modelling text in a wide range of languages.
The abundance of textual data in different languages across the internet has fueled the progress of multi-lingual models \cite{johnson2017google,xue2020mt5,siddhant2022towards}.
On the other hand, speech technologies are prevalent in smartphones and personal assistants, but their language availability is relatively limited compared to the languages that LLMs support \cite{baevski2020wav2vec, radford2023robust}.

\definecolor{GREEN-ONE}{RGB}{194,230,153}
\definecolor{GREEN-TWO}{RGB}{161,217,155}
\definecolor{GREEN-THREE}{RGB}{49,163,84}
\definecolor{GREEN-FOUR}{RGB}{35,132,67}

\begin{figure}[t]
    \centering
    \includegraphics{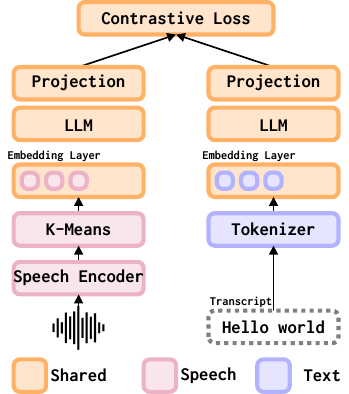}
    \caption{
    \textbf{Our dual encoder architecture and training pipeline}. 
    We expand the embedding layer of our backbone LLM to support the additional discretized speech tokens, that are extracted from a pre-trained speech encoder.
    At the same time, we tokenize the corresponding transcripts with the LLM tokenizer.
    We encode the speech tokens and transcripts separately and train the model with a contrastive loss over the dot product between speech and transcript embeddings. 
    }
    \label{fig:architecture}
\end{figure}

Various efforts have explored solutions to the speech-text data scarcity problem \cite{duquenne2021multimodal, ardila2019common, wang2020covost}. 
Works such as SpeechMatrix \cite{duquenne2022speechmatrix} use separate speech and text encoders to mine semantically similar utterances that are neighbors in an embedding space.  
However, these approaches are limiting because they require speech and text encoders that have aligned representation spaces. 

We posit that we can retrieve speech and text utterances by aligning both modalities within the embedding space built from a single pre-trained LLM. 
We take inspiration from previous works that use pre-trained LLMs to perform automatic speech recognition (ASR) and automatic speech translation (AST) \cite{rubenstein2023audiopalm, wang2023neural, hassid2023textually, gong2023joint, peng2023prompting}. 
Our intuition is that we can perform the speech and text alignment leveraging the capabilities of text-only LLMs without requiring two separate models.  

In this paper, we propose converting LLMs into speech and text DE retrieval systems without requiring speech pre-training and outperform previous methods with significantly less data. 
By discretizing speech into acoustic units \cite{hsu2021hubert}, we extend our LLMs embedding layer and treat the acoustic units as ordinary text tokens.
Consequently, we transform our LLM into a retrieval system via a contrastive loss allowing us to match speech and text utterances in various languages.
Our contributions are the following: 
\begin{enumerate}[itemsep=1pt,topsep=2pt,parsep=2pt]
    \item We build a speech-to-text symmetric DE from a pre-trained LLM. We show that our retrieval system is effective matching speech and text in 102 languages of FLEURS \cite{conneau2023fleurs} despite only training on 21 languages.  
    \item We show that our model exhibits cross-lingual speech and text matching without training on this type of data. At the same time, we find that cross-lingual speech and text matching is further improved by training on readily available machine translation data. 
\end{enumerate}

\section{Method}
We train a transformer-based DE model that encodes speech and text given a dataset $\emph{D} = \{(x_i, y_i)\}$, where $x_i$ is a speech utterance and $y_i$ is its transcription.
We denote the speech and text embeddings as $\bm{x_i} = E(x_i)$ and $\bm{y_i} = E(y_i)$, respectively, where $E$ is a transformer-based DE that encodes speech and text.

\subsection{Generating Audio Tokens}

We convert raw speech into discrete tokens using the process in \citet{lakhotia-etal-2021-generative,borsos2023audiolm}.
The process converts a speech query $x_i$ into an embedding using a pre-trained speech encoder.
The output embedding is then discretized into a set of tokens using k-means clustering.
We refer to the resulting tokens as {\it audio tokens}. 
We use the 2B variant of the Universal Speech Model (USM) encoder \cite{zhang2023google} as the speech encoder and take the middle layer as the embedding for $x_i$.
Additionally, we generate audio tokens at 25Hz using k-means clustering \footnote{We use the \textbf{USM-v2} audio tokenizer from \citet{rubenstein2023audiopalm}}.
We will refer to this as our {\it audio token vocabulary}.  

\subsection{Supporting Text and Audio Tokens}

To support text and audio tokens in our LLM, we follow the formulation of \citet{rubenstein2023audiopalm}.
We extend the embedding layer of a transformer decoder by $a$ tokens, where $a$ represents the size of our audio token vocabulary. 
This modification leads to an embedding layer with size $(t + a) \times m$, where $t$ is the number of tokens in the text vocabulary and $m$ is the dimensions of the embedding vectors.
In our implementation, the first $t$ tokens represent text and the remaining $a$ tokens are reserved for audio. 
We initialize the embeddings layer from scratch when training our model. 

\section{Data and Tasks}

Appendix \ref{sec:dataset_stats} details our training and evaluation datasets along with the number of languages in each dataset, the split we used, and the size of each dataset. 
We focus on the following retrieval tasks:

\paragraph{Speech-to-Text Retrieval (S2T)} 
involves retrieving the corresponding transcription from a database given a speech sample.
In S2T, we train on CoVoST-2 \cite{Wang2021CoVoST2A} speech utterances and their transcriptions.
CoVoST-2 is a large multilingual speech corpus derived from Wikipedia expanding over 21 languages and provides translation to and from English. 
We use FLEURS \cite{conneau2023fleurs} to evaluate S2T performance on 102 languages.
FLEURS is an $n$-way parallel dataset containing speech utterances from FLoRES-101 \cite{Goyal2021TheFE} human translations.
To evaluate S2T, we report recall at 1 ($R@1$) rates for retrieving the correct transcription for every speech sample and word error rate (WER). 

\paragraph{Speech-to-Text Translation Retrieval (S2TT)} attempts to retrieve the corresponding text translation of a speech sample.
We use S2TT to measure the cross-lingual capabilities of our multi-modal DE retrieval system. 
We evaluate this capability zero-shot on X $\to$ En S2TT data of FLUERS and explore if we can further improve this capability by training on readily-available machine translation data from WikiMatrix \cite{Schwenk2019WikiMatrixM1}.
We pick French, German, Dutch, and Polish to English that are common across WikiMatrix and FLEURS and further discuss the amount of machine translation data used in Appendix \ref{sec:dataset_stats}.
For S2TT, we report 4-gram corpusBLEU \cite{ post-2018-call}.

\section{Model}

Figure \ref{fig:architecture} shows an illustration of our model.
We initialize our dual encoder from PaLM 2 XXS \cite{anil2023palm} and append a linear projection layer after pooling the outputs along the sequence length dimension.
The embedding and linear projection layers are initialized randomly. 
After initializing our model from PaLM 2, we use a contrastive loss \cite{Hadsell2006DimensionalityRB}. 
Appendix \ref{sec:training_setup} includes more details on our training setup. 
We will refer to our proposed model as PaLM 2 DE.

\section{Experiments}

\begin{table}[t]
\centering
\resizebox {\linewidth} {!} {
\begin{tabular}{lrr}
\toprule
 & \multicolumn{1}{l}{\emph{R@1} $\uparrow$} & \multicolumn{1}{l}{\emph{WER} $\downarrow$} \\ \midrule
mSLAM DE \cite{conneau2023fleurs} & 76.9 & 14.6 \\ 
PaLM 2 DE (Proposed Model) & 86.7 & 13.4 \\
\bottomrule
\end{tabular}
}
\caption{PaLM 2 DE results for \emph{R@1} and WER compared against the mSLAM DE on 102 languages from FLEURS for speech-to-text retrieval (S2T).}
\label{tab:speech_to_text_retrival_baseline_comparison}
\end{table}

We train our DE model to perform S2T, where the task is to retrieve the corresponding transcription given a speech sample.
We train on the 21 languages from CoVoST-2 and evaluate our model using the S2T portion of FLEURS in 102 languages.

\subsection{Speech-to-Text Retrieval}

Table \ref{tab:speech_to_text_retrival_baseline_comparison} shows the average \emph{R@1} and WER for S2T for 102 languages from FLEURS.
We compare against the mSLAM DE model from \citet{conneau2023fleurs}, a model trained on 426k hours of S2T data in 51 languages and fine-tuned on FLEURS training data.  
Our model significantly outperforms the mSLAM DE baseline in \emph{R@1} and $WER$ metrics despite being trained with only 1/10 of the data and having been initialized from a text-only LLM.
More importantly, our model was only trained on the 21 languages in CoVoST-2 and never fine-tuned on the FLEURS training data.

\subsubsection{Seen-Unseen Breakdown}

\definecolor{GREEN-ONE}{RGB}{240,249,232}
\definecolor{GREEN-TWO}{RGB}{186,228,188}
\definecolor{GREEN-THREE}{RGB}{123,204,196}
\definecolor{GREEN-FOUR}{RGB}{43,140,190}

\begin{figure}[t]
  \centering
  \includegraphics{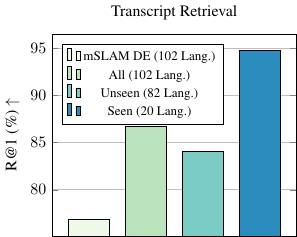}
  \caption{\emph{R@1} transcription retrieval for seen and unseen languages in the training set.}
  \label{fig:seen_vs_unseen_r_1}
\end{figure}

In Figure \ref{fig:seen_vs_unseen_r_1} we break down the \emph{R@1} scores based on seen and unseen languages during training.
We find that our model performs best on the 20 languages that are within the training and evaluation data, but still perform well on the remaining 82 unseen languages.
We hypothesize this is due to the vast textual multilingual data our backbone LLM has seen during pre-training. 

\subsubsection{Language Group Breakdown}

\begin{table}[t]
  \centering
\resizebox {\linewidth} {!} {
  \begin{tabular}{lrr|r}
    \toprule
    &\multicolumn{2}{c}{\textit{R@1} $\uparrow$} &  \\
    \cmidrule{2-3}
    {Language Group (\#)} & {mSLAM DE} & {PaLM 2 DE} & \multicolumn{1}{c}{\multirow{2}{*}{$\#$ Wins}} \\
    & { \cite{conneau2023fleurs}} & {(Proposed Model)} & \\
    \midrule
    Afro-Asiatic (7) & 73.67 & \textbf{84.22} & 5 \\
    Atlantic-Congo (14) & \textbf{86.77} & 70.41 & 1 \\
    Austro-Asiatic (2) & \textbf{47.90} & 34.42 & 0\\
    Austronesian (6) & 75.50 & \textbf{90.73}  & 6 \\
    Dravidian (4) & 65.70 & \textbf{92.06} & 4 \\
    Indo-European (51) & 84.62 & \textbf{95.32} & 49 \\
    Japonic (1) & 5.80 & \textbf{91.54} & 1 \\
    Kartvelian (1) & 70.50 & \textbf{82.92} & 1 \\
    Koreanic (1) & 5.20 & \textbf{52.36} & 1 \\
    Kra-Dai (2) & 3.20 & \textbf{22.09} & 1 \\
    Mongolic (1) & 70.70 & \textbf{99.89} & 1 \\
    Nilo-Saharan (1) & 91.00 & \textbf{92.52} & 1 \\
    Sino-Tibetan (3) & 3.40 & \textbf{90.66} & 3 \\
    Turkic (5) & 81.28 & \textbf{92.86} & 4 \\
    Uralic (3) & 91.40 & \textbf{99.04} & 3 \\
    \midrule
    All (102) & 76.90        & \textbf{86.72} & 81 \\
    \bottomrule
  \end{tabular}
  }
  \caption{FLEURS S2T (\emph{R@1}) performance by language groups. Bold represents better performance. Numbers in parenthesis are the number of languages within the language group. $\#$ Wins is the number of languages where PaLM 2 DE outperforms mSLAM in the language group.}
  \label{tab:language_group_breakdown_r_1}
\end{table}

Table \ref{tab:language_group_breakdown_r_1} shows the \emph{R@1} language group breakdown for S2T on FLEURS. 
We find that although we only trained on 21 languages, our model significantly outperforms mSLAM DE in 13 of the 15 language groups.
These results are consistent with the experiments in \citet{hassid2023textually} which explore the effect of initializing speech language models from pre-trained LLMs.

\subsection{Evaluating on Cross-Modal and Cross-Lingual Tasks}

\definecolor{GREEN-ONE}{RGB}{240,249,232}
\definecolor{GREEN-TWO}{RGB}{186,228,188}
\definecolor{GREEN-THREE}{RGB}{123,204,196}
\definecolor{GREEN-FOUR}{RGB}{43,140,190}

\begin{figure}[t]
  \centering
  \includegraphics{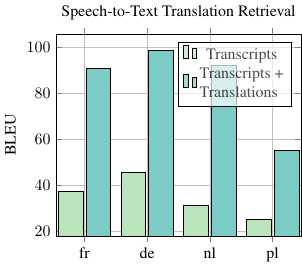}
  \caption{BLEU scores for FLEURS zero-shot S2TT when training on \texttt{\textbf{Transcripts}} or \texttt{\textbf{Transcripts + Translations}} for PaLM 2 DE. Combining transcripts and translation data improves zero-shot S2TT retrieval.}
  \label{fig:cv_vs_cv+wm_bleu}
\end{figure}

We evaluate on S2TT to gauge the cross-modal and cross-lingual capabilities of our model.
We show we can improve S2TT by simply combining S2T and translation data without S2TT training data.

\subsubsection{Zero-Shot S2TT}
Given the multi-lingual capabilities of our backbone language model, we explore if these capabilities are transferred after training our model contrastively on the S2T task. 
We hypothesize that our model should showcase cross-lingual and cross-modal capabilities due to the cross-modal training task and the cross-lingual capabilities of the backbone LLM.
We evaluate S2TT in a zero-shot setting to assess our model's performance retrieving English translations given a speech sample in another language. 
Using the FLEURS S2TT portion, we evaluate S2TT X $\to$ En in 4 languages: German, Polish, French, and Dutch.

Figure \ref{fig:cv_vs_cv+wm_bleu} shows BLEU S2TT performance using S2T CoVoST-2 in 21 languages. 
We call this setup \texttt{\textbf{Transcripts}} in Figure \ref{fig:cv_vs_cv+wm_bleu}.
Our results demonstrate that even when only training our model on speech and transcriptions, we can achieve some zero-shot S2TT performance and We find that S2TT BLEU scores are considerably higher for languages present S2T training data.
For example, Polish was not in the S2T training therefore its BLEU scores are the lowest. 

\subsubsection{Improving S2TT with MT Data}

To further improve our model's cross-lingual performance, we add readily available translation data from \citet{Schwenk2019WikiMatrixM1} to improve S2TT. 
For each batch, we combine 25\% translation and 75\% S2T data.
Figure \ref{fig:cv_vs_cv+wm_bleu} shows comparison of only training on S2T (\textbf{\texttt{Transcripts}}) and combining S2T and translation data (\textbf{ \texttt{Transcriptions + Translations}}).
We find that combining S2T and translation data significantly improves the S2TT BLEU scores in all 4 languages without training on S2TT data.
This finding demonstrates that we can improve our models cross-lingual performance with highly accessible translation data without needing scarce and often expensive speech-to-text translation training data.

\section{Related Work}

The success of pre-trained LLMs have motivated the application of these models in different modalities. \citet{lakhotia-etal-2021-generative} transformed speech into pseudo-text units to introduce the task of generative spoken language modeling. 
\citet{borsos2023audiolm} introduced a framework to generate audio with long-term consistency. Consequently, \citet{hassid2023textually} showed that SpeechLMs benefit from being initialized from pre-train LLMs while \citet{rubenstein2023audiopalm} demonstrated that pre-trained LLMs can be adapted to various tasks that required text and speech understanding.

On the other hand, several works aim to build joint speech and text representations \cite{khurana2022samu, gow-smith-etal-2023-naver}.
\citet{chung2021w2v} introduced w2v-bert which combines masked language modeling and contrastive learning to create speech representations. 
\citet{bapna2022mslam} jointly pre-trains on speech and text from unsupervised speech and text data.
Recently, \citet{duquenne2023sentence} employed separate speech and text encoders to generate embeddings in over 200 languages. 
Nevertheless, there is still a lack of understanding of whether joint speech and text representations can be built from a single encoder.
We fill this gap by using pre-trained LLMs to jointly train on speech samples and their transcriptions to show that our approach is capable of speech-text matching in 102 languages.

\section{Conclusion}

We present an effective approach to developing a speech-to-text DE from a text-only LLM. 
Our findings suggest that by using a text-only LLM as a backbone model, we can drastically outperform previous approaches using considerably less speech-to-text training data.
Additionally, we find that we can improve zero-shot speech translation by simply combining readily available translation and S2T data. 
We showcase our findings in 102 languages for S2T and 4 languages in S2TT; opening up the possibility of using speech-to-text DE's in different cross-model and cross-lingual settings.

\section{Acknowledgements}

We would like to thank Shankar Kumar, Ankur Bapna, and the anonymous reviewers for the valuable feedback on the draft of the paper. 
Chris Tar, Mario Guajardo-Céspedes, and Jason Riesa for the early experiment discussions and feedback.
Christian Frank, Duc Dung Nguyen, Alex Tudor, and Dalia El Badawy for helping answer questions about AudioPaLM.

\bibliography{custom}

\appendix

\section{Appendix}
\label{sec:appendix}

\begin{table*}[t]
    \centering
      \resizebox {\linewidth} {!} {

    \begin{tabular}{l|r|r}
        \toprule
        Input Type & Before Tokenization & Input Ids \\
        \midrule
        Speech  & \texttt{[English Speech]\textbf{ 50,210,245,} $\ldots$} & \texttt{240, 503, \textbf{32050, 32210, 32245,} $\ldots$} \\
        Transcription & \texttt{[English Text] Hello World .} & \texttt{59, 294, 691, $\ldots$} \\
        \bottomrule
    \end{tabular}
    }
    \caption{Example of the speech and transcript inputs given to our model. The speech input is composed of a prefix containing the language and the input modality. Text will be tokenized using the LLMs tokenizer and an offset will be applied to the audio token to match the tokens that were reserved within the audio token vocabulary. Bold numbers represent the audio tokens before tokenization and after the offset is applied to the audio tokens.}
    \label{tab:tokenization}
\end{table*}

\subsection{Training Setup}
\label{sec:training_setup}

\citet{ni-etal-2022-sentence} showed that applying a contrastive loss to sentence encoders leads to improved retrieval performance in downstream tasks. 
After initializing our model from the PaLM 2, we use a contrastive loss \cite{Hadsell2006DimensionalityRB}. 

\begin{equation}
L = -\frac{1}{N} \sum_{i=1}^{N} \frac{e^{\text{sim}(\bm{x}_{i}, \bm{y}_{i})}}{\sum_{j=1}^{N} e^{\text{sim}(\bm{x}_{i}, \bm{y}_{j})}}\label{eq:loss}
\end{equation}

Using equation \ref{eq:loss}, our multi-modal DE will learn from paired speech and text embeddings $(\bm{x}_i, \bm{y}_i)$, where $\bm{y}_{i}$ is considered as a positive example to $\bm{x}_i$ while all other examples where $i \neq j$ are negative ones.
The model should learn to bring the positive transcriptions closer to the corresponding speech sample, while pushing away all the other negative transcriptions. 
In our training, the positive and negative distinction is done within the training batch. 
Hence, we apply an in-batch softmax as part of our loss computation.
Lastly, \textit{sim()} is a similarity function formulated as the dot product between the speech sample and the transcription embeddings.

To train our model, we use the sum of a contrastive loss with a spreadout loss \cite{Zhang2017LearningSL} of both the speech and text embeddings. 
We calculate the contrastive loss \cite{yang2019improving} in a bidirectional way, by adding the loss in the speech-to-text and the text-to-speech direction. 

We use the Adam \cite{Kingma2014AdamAM} optimizer with a learning rate of $1.0 \times 10^{-3}$ with linear ramp cosine decay scheduler with 2.5k warm up steps.
We use a dropout probability of $0.1$ and train for 100k steps with a batch size of 1024.

\subsection{Expressing Tasks}
\label{sec:expressing_tasks}
For training and inference, we found that using a prefix improves speech-to-text retrieval performance.
Therefore, we pre-pend a prefix containing the language and modality shown in in Table \ref{tab:tokenization}.
In the case of a speech utterance, the prefix will be tokenized with the LLMs tokenizer and the remaining will be converted to audio tokens. 

\subsection{Data}
\begin{table}[t]
  \centering
  \resizebox {\linewidth} {!} {

  \begin{tabular}{l l l r l r}
    \toprule
    Dataset & Type & Task & Langs. & Split & Size \\
    \midrule
    CoVoST-2 &  Speech & S2T & 21 & Train & 900 h. \\
    FLEURS &  Speech & S2T & 102 & Test & 283 h. \\
    FLEURS & Speech & S2TT & 102 & Test & 283 h. \\
    \midrule 
    Wikimatrix & Text & MT & 4 & Train & 9M sents. \\
    \bottomrule
  \end{tabular}
  }
  \caption{Training and evaluation datasets. CoVoST-2 is used for speech-to-text  retrieval (S2T), Wikimatrix is for machine translation retrieval (MT), and FLEURS is for evaluating  X $\to$ En  speech-to-text translation retrieval (S2TT) and also speech-to-text retrieval (S2T).}
  \label{tab:dataset_stats}
\end{table}
\begin{table}[t]
\centering
\begin{tabular}{lr}
\toprule
 & \multicolumn{1}{l}{\emph{\# Sents. X $\to$ En}} \\ \midrule
German (de) & 6.2M \\
Polish (pl) & 2.1M \\
French (fr) & 705k \\
Dutch (nl) & 570k \\ 
\bottomrule
\end{tabular}
\caption{Number of parallel sentences used in the machine translation mixture from Wikimatrix corpus.}
\label{tab:wikimatrix_sentences}
\end{table}
\label{sec:dataset_stats}

Table \ref{tab:dataset_stats} shows the training and evaluation datasets we used through out our experiments.
We used 21 languages CoVoST-2 to train our model on speech-to-text retrieval which amounts to approximately 900 hours of speech.
To evaluate our models speech-to-text retrieval capabilities, we evaluate on FLEURS speech-to-text test split on 102 languages.
We use FLEURS speech-to-text translation test split to evaluate our models abilities on tasks that require cross-lingual and cross-modal knowledge.
We evaluate of 4 different languages: German, Polish, French, and Dutch.

We find that combining speech-to-text retrieval data and readily available translation data improves our models cross-lingual and cross-modal abilities. 
Table \ref{tab:wikimatrix_sentences} shows the number of parallel sentences we used during training from X $\to$ En. 




\subsection{Performance Breakdown By Language}

Table \ref{tab:fleurs_recall_per_language} includes the PaLM 2 DE \emph{R@1} for each language found in FLEURS.
We also include the language group from Table \ref{tab:language_group_breakdown_r_1} and the number of examples found within each S2T test set.

\clearpage


\begin{table*}[ht!]
\centering
\resizebox {\linewidth} {!} {
\begin{tabular}[ht!]{l|lllr|rr}
\toprule
Idx & Language Name                             & Code          & Family        & $\#$ Examples &  \multicolumn{2}{c}{\emph{R@1}} \\
\cmidrule{1-7}
    &                                           &               &               &          &                 mSLAM & PaLM 2 DE \\
\midrule

1   & Afrikaans      & af  & Indo-European  & 414  & 90.1 & 99.3  \\
2   & Amharic        & am  & Afro-Asiatic   & 516  & 34.1 & 69.6  \\
3   & Arabic         & ar  & Afro-Asiatic   & 427  & 82.7 & 98.8  \\
4   & Armenian       & hy  & Indo-European  & 929  & 50.3 & 89.7  \\
5   & Assamese       & as  & Indo-European  & 980  & 81.5 & 87.4  \\
6   & Asturian       & ast & Indo-European  & 946  & 90.1 & 100.0 \\
7   & Azerbaijani    & az  & Turkic         & 918  & 83.0 & 98.4  \\
8   & Belarusian     & be  & Indo-European  & 955  & 90.2 & 97.2  \\
9   & Bengali        & bn  & Indo-European  & 911  & 83.5 & 84.6  \\
10  & Bosnian        & bs  & Indo-European  & 923  & 95.5 & 99.8  \\
11  & Bulgarian      & bg  & Indo-European  & 657  & 95.1 & 100.0 \\
12  & Burmese        & my  & Sino-Tibetan   & 870  & 2.4  & 19.3  \\
13  & Cantonese      & yue & Sino-Tibetan   & 819  & 2.4  & 83.6  \\
14  & Catalan        & ca  & Indo-European  & 938  & 93.2 & 100.0 \\
15  & Cebuano        & ceb & Austronesian   & 532  & 79.8 & 94.9  \\
16  & Croatian       & hr  & Indo-European  & 914  & 98.0 & 99.8  \\
17  & Czech          & cs  & Indo-European  & 720  & 98.1 & 99.6  \\
18  & Danish         & da  & Indo-European  & 929  & 94.1 & 99.9  \\
19  & Dutch          & nl  & Indo-European  & 364  & 95.3 & 100.0 \\
20  & English        & en  & Indo-European  & 647  & 96.0 & 99.1  \\
21  & Estonian       & et  & Uralic         & 892  & 95.6 & 99.9  \\
22  & Filipino       & fil & Austronesian   & 928  & 73.1 & 89.1  \\
23  & Finnish        & fi  & Uralic         & 916  & 93.0 & 98.9  \\
24  & French         & fr  & Indo-European  & 675  & 90.7 & 100.0 \\
25  & Fula           & ff  & Atlantic-Congo & 649  & 81.4 & 81.7  \\
26  & Galician       & gl  & Indo-European  & 927  & 90.9 & 100.0 \\
27  & Ganda          & lg  & Atlantic-Congo & 705  & 90.7 & 75.7  \\
28  & Georgian       & ka  & Kartvelian     & 978  & 70.5 & 82.9  \\
29  & German         & de  & Indo-European  & 841  & 91.2 & 100.0 \\
30  & Greek          & el  & Indo-European  & 649  & 81.2 & 73.2  \\
31  & Gujarati       & gu  & Indo-European  & 1000 & 77.0 & 95.9  \\
32  & Hausa          & ha  & Afro-Asiatic   & 557  & 84.5 & 83.1  \\
33  & Hebrew         & he  & Afro-Asiatic   & 792  & 64.0 & 76.0  \\
34  & Hindi          & hi  & Indo-European  & 417  & 78.0 & 83.7  \\
35  & Hungarian      & hu  & Uralic         & 902  & 85.3 & 98.3  \\
36  & Icelandic      & is  & Indo-European  & 46   & 71.7 & 97.8  \\
37  & Igbo           & ig  & Atlantic-Congo & 869  & 85.8 & 64.9  \\
38  & Indonesian     & id  & Austronesian   & 684  & 79.6 & 99.4  \\
39  & Irish          & ga  & Indo-European  & 829  & 55.1 & 69.5  \\
40  & Italian        & it  & Indo-European  & 857  & 93.5 & 100.0 \\
41  & Japanese       & ja  & Japonic        & 650  & 5.8  & 91.5  \\
42  & Javanese       & jv  & Austronesian   & 722  & 78.0 & 97.0  \\
43  & Kabuverdianu   & kea & Indo-European  & 859  & 95.4 & 99.9  \\
\bottomrule
  \end{tabular}
}
\end{table*}


\begin{table*}[ht!]
\centering
\resizebox {\linewidth} {!} {
\begin{tabular}[ht!]{l|lllr|rr}
\toprule
Idx & Language Name                             & Code          & Family        & $\#$ Examples &  \multicolumn{2}{c}{\emph{R@1}} \\
\cmidrule{1-7}
    &                                           &               &               &          &                 mSLAM & PaLM 2 DE \\
\midrule
44  & Kamba          & kam & Atlantic-Congo & 798  & 89.7 & 81.5  \\
45  & Kannada        & kn  & Dravidian      & 831  & 69.0 & 88.8  \\
46  & Kazakh         & kk  & Turkic         & 841  & 88.7 & 83.1  \\
47  & Khmer          & km  & Austro-Asiatic & 765  & 42.1 & 20.3  \\
48  & Korean         & ko  & Koreanic       & 382  & 5.2  & 52.4  \\
49  & Kyrgyz         & ky  & Turkic         & 974  & 84.3 & 88.6  \\
50  & Lao            & lo  & Kra-Dai        & 399  & 37.0 & 23.3  \\
51  & Latvian        & lv  & Indo-European  & 848  & 97.4 & 100.0 \\
52  & Lingala        & ln  & Atlantic-Congo & 440  & 91.2 & 76.4  \\
53  & Lithuanian     & lt  & Indo-European  & 985  & 96.8 & 98.2  \\
54  & Luo            & luo & Nilo-Saharan   & 254  & 91.0 & 92.5  \\
55  & Luxembourgish  & lb  & Indo-European  & 929  & 80.5 & 74.6  \\
56  & Macedonian     & mk  & Indo-European  & 967  & 96.1 & 98.8  \\
57  & Malay          & ms  & Austronesian   & 749  & 77.7 & 98.7  \\
58  & Malayalam      & ml  & Dravidian      & 944  & 62.3 & 88.3  \\
59  & Maltese        & mt  & Afro-Asiatic   & 918  & 92.7 & 76.0  \\
60  & Mandarin       & cmn & Sino-Tibetan   & 944  & 5.4  & 100.0 \\
61  & Maori          & mi  & Austronesian   & 890  & 64.7 & 65.3  \\
62  & Marathi        & mr  & Indo-European  & 1005 & 69.8 & 82.4  \\
63  & Mongolian      & mn  & Mongolic       & 949  & 70.7 & 99.9  \\
64  & Nepali         & ne  & Indo-European  & 724  & 66.1 & 89.6  \\
65  & Northern-Sotho & nso & Atlantic-Congo & 738  & 80.8 & 70.3  \\
66  & Norwegian      & nb  & Indo-European  & 357  & 91.9 & 100.0 \\
67  & Nyanja         & ny  & Atlantic-Congo & 745  & 85.5 & 63.6  \\
68  & Occitan        & oc  & Indo-European  & 968  & 77.4 & 99.4  \\
69  & Oriya          & or  & Indo-European  & 875  & 15.7 & 95.1  \\
70  & Oromo          & om  & Afro-Asiatic   & 41   & 92.7 & 100.0 \\
71  & Pashto         & ps  & Indo-European  & 510  & 84.8 & 91.0  \\
72  & Persian        & fa  & Indo-European  & 858  & 85.4 & 100.0 \\
73  & Polish         & pl  & Indo-European  & 758  & 95.8 & 99.3  \\
74  & Portuguese     & pt  & Indo-European  & 914  & 91.9 & 99.9  \\
75  & Punjabi        & pa  & Indo-European  & 574  & 70.6 & 96.7  \\
76  & Romanian       & ro  & Indo-European  & 882  & 92.0 & 100.0 \\
77  & Russian        & ru  & Indo-European  & 774  & 93.2 & 100.0 \\
78  & Serbian        & sr  & Indo-European  & 700  & 97.7 & 99.1  \\
79  & Shona          & sn  & Atlantic-Congo & 920  & 84.1 & 53.9  \\
80  & Sindhi         & sd  & Indo-European  & 977  & 71.8 & 85.4  \\
81  & Slovak         & sk  & Indo-European  & 791  & 97.6 & 99.5  \\
82  & Slovenian      & sl  & Indo-European  & 834  & 97.4 & 100.0 \\
83  & Somali         & so  & Afro-Asiatic   & 1007 & 68.7 & 86.0  \\
84  & Sorani-Kurdish & ckb & Indo-European  & 918  & 80.8 & 96.7  \\
85  & Spanish        & es  & Indo-European  & 907  & 69.6 & 100.0 \\
86  & Swahili        & sw  & Atlantic-Congo & 487  & 91.2 & 86.2  \\
\bottomrule
  \end{tabular}
}
\end{table*}

\begin{table*}[ht!]
\centering
\resizebox {\linewidth} {!} {
\begin{tabular}[t]{l|lllr|rr}
\toprule
Idx & Language Name                             & Code          & Family        & $\#$ Examples &  \multicolumn{2}{c}{\emph{R@1}} \\
\cmidrule{1-7}
    &                                           &               &               &          &                 mSLAM & PaLM 2 DE \\
\midrule
87  & Swedish        & sv  & Indo-European  & 758  & 94.2 & 100.0 \\
88  & Tajik          & tg  & Indo-European  & 590  & 76.3 & 92.7  \\
89  & Tamil          & ta  & Dravidian      & 582  & 58.0 & 98.1  \\
90  & Telugu         & te  & Dravidian      & 471  & 73.5 & 93.0  \\
91  & Thai           & th  & Kra-Dai        & 1011 & 3.2  & 20.9  \\
92  & Turkish        & tr  & Turkic         & 742  & 84.5 & 100.0 \\
93  & Ukrainian      & uk  & Indo-European  & 750  & 93.5 & 99.3  \\
94  & Umbundu        & umb & Atlantic-Congo & 264  & 77.3 & 62.1  \\
95  & Urdu           & ur  & Indo-European  & 299  & 70.6 & 91.3  \\
96  & Uzbek          & uz  & Turkic         & 861  & 67.6 & 94.2  \\
97  & Vietnamese     & vi  & Austro-Asiatic & 850  & 64.5 & 48.6  \\
98  & Welsh          & cy  & Indo-European  & 1002 & 82.3 & 96.1  \\
99  & Wolof          & wo  & Atlantic-Congo & 351  & 90.6 & 87.5  \\
100 & Xhosa          & xh  & Atlantic-Congo & 1034 & 90.9 & 30.2  \\
101 & Yoruba         & yo  & Atlantic-Congo & 816  & 92.4 & 84.6  \\
102 & Zulu           & zu  & Atlantic-Congo & 822  & 85.5 & 67.2  \\ \bottomrule
    & All  (102)       &     &                &      & 76.9 & 86.7  \\
\bottomrule
  \end{tabular}
}
\caption{Language name, code, family, and number of examples for each test set found in FLEURS. We report \emph{R@1} for mSLAM and PaLM 2 DE.}
\label{tab:fleurs_recall_per_language}
\end{table*}

\end{document}